\documentclass[runningheads]{llncs}
\usepackage[T1]{fontenc}
\usepackage{listings}
\usepackage{mdframed}
\usepackage{caption}
\usepackage{multirow}
\usepackage{multicol}
\newcommand\Tstrut{\rule{0pt}{2.6ex}}         % = `top' strut
\newcommand\Bstrut{\rule[-0.9ex]{0pt}{0pt}}   % = `bottom' strut
\newcommand{\para}[1]{\paragraph{\textnormal{\textbf{#1.}}}}

\usepackage{graphicx}
\usepackage{appendix}

\begin{document}
\title{Fighting Fire with Fire: Adversarial Prompting to Generate a Misinformation Detection Dataset}
\titlerunning{Adversarial Prompting to Generate a Misinformation Detection Dataset}

\author{Shrey Satapara\inst{1}\orcidID{0000-0001-6222-1288}
\and
Parth Mehta\inst{2}\orcidID{0000-0002-4509-1298}
 \and
Debasis Ganguly \inst{3}\orcidID{0000-0003-0050-7138} \and
Sandip Modha \inst{4}\orcidID{0000-0003-2427-2433}
}

\authorrunning{S. Satapara et al.}
\institute{IIT Hyderabad, India\\
%\email{shreysatapara@gmail.com}
\and
Parmonic, USA\\
%\email{parth.mehta126@gmail.com}
\and
University of Glasgow, UK\\
%\email{debforit@gmail.com}
\and LDRP Institute of Technology and Research, India\\
%\email{sjmodha@gmail.com}
}

\maketitle              % typeset the header of the contribution

\begin{abstract}
The recent success in language generation capabilities of large language models (LLMs), such as GPT, Bard, Llama etc., can potentially lead to concerns about their possible misuse in inducing mass agitation and communal hatred via  
generating fake news and spreading misinformation. Traditional means of developing a misinformation ground-truth dataset does not scale well because of the extensive manual effort required to annotate the data. In this paper, we propose an LLM-based approach of creating silver-standard ground-truth datasets for identifying misinformation.
Specifically speaking, given a trusted news article, 
our proposed approach involves prompting LLMs to automatically generate a summarised version of the original article. The prompts in our proposed approach act as a controlling mechanism to generate specific types of factual incorrectness in the generated summaries, e.g., incorrect quantities, false attributions etc.
%
%process caters to a number of different types of misinformation including incorrect quantities, false attributions etc.
%We conducted a manual review of a small subset of 20 summaries from each category from this synthetically generated data for quality checking, and observed that the generated data in most cases correspond to factually incorrect information of a desired specific type, e.g., false attribution.
To investigate the usefulness of this dataset, we conduct a set of experiments where we train a range of supervised models for the task of misinformation detection.

%The work not only highlights the limitations of the checks that these LLMs have in place to prevent their misuse but also shows a way to exploit these limitations in a positive manner. Such limitations can be used to fight the very issues that could arise from their misuse. We use adversarial prompting to generate summaries of news articles with incorrect and misleading information. Such a dataset will not only aid in fact-checking research but also improve explainability.

\end{abstract}
\section{Introduction}

Until recently news articles were one of the most dominant sources of information about a wide variety of topics and temporal events. However, as access to the internet proliferates, a large number of relatively novice and untrustworthy information sources have mushroomed around us. As a result, news has lost its traditional value of reliability and is no longer limited to established media houses or reputed sources. Virtually anyone can put up a website with information on a topic of their choice and easily disseminate it by leveraging the power of social media. While this has democratized access to knowledge, it has also fostered the rise of misinformation and fake news. This phenomenon is only amplified by social media echo chambers that have demonstrated the ability to shape and polarize public opinion.

The rapid advancement of language generation models, particularly large language models (LLMs) such as GPT \cite{openai2023gpt4}, Llama \cite{touvron2023llama}, Bard \cite{manyika2023overview}, etc. has further intensified this issue. These models can generate highly convincing yet potentially misleading content, which can happen both unintentionally due to the inherent limitations of LLMs, commonly known as hallucinations \cite{zhang2023hallucination}, or can be done deliberately by miscreants via adversarial prompting \cite{zou2023adverserial}. All LLMs are prone to hallucinations to varying extents, where they generate factually incorrect or misleading information unintentionally and unknowingly. LLM-generated text is usually fluent and hence can potentially appear to be of adequate enough quality to instil trust among readers. On the other hand, miscreants can also leverage the powers of LLMs to intentionally generate false information, ranging from introducing a subtle bias to creating completely fabricated narratives that suit their purpose. While most LLMs claim to have safeguards in place, adversarial prompting or jailbreaking is a common phenomenon and it is indeed possible to intentionally generate misleading content using these LLMs, as we show in this paper. Such content can then be widely circulated leading to potential law and order issues.

Detecting fake or misguiding information is not a novel area and has been actively researched for several years now. However, traditional fact-checking mechanisms depend on validating such content against reliable information from verified sources often with substantial human efforts. The possibility of using LLMs for generating misleading or fake content has added a new dimension to the already complex problem. With the available technology, it is now possible to generate massive quantities of factually incorrect content with targeted ulterior motives in a very short span of time. The verification mechanism, on the other hand, remains slow and mostly human-dependent. New approaches that are less dependent on human involvement or annotations and can scale up quickly are required to combat such a threat.

\para{Relevance of our work to the `IR for social good' track}
To address this concern of rapid misinformation propagation (which causes societal harms, such as communal hatred, mass agitations etc.), it is imperative to develop robust mechanisms that can effectively distinguish between authentic and fake content, which makes this work relevant to the `IR for social good' track.

%as well as identify machine-generated content from human written content.

The first step towards misinformation detection is often the creation of a dataset containing both genuine as well as misleading content. Such datasets aid researchers in training their models as well as provide a common benchmark for comparing the results. However, creating such a dataset requires significant effort and time. Traditionally this is done by collecting data from fact-checking websites \cite{shu2020fakenewsnet,shu2017fake,wang2017liar}. This is a resource-consuming effort and also dependent on the existence of such fact-checking websites in the first place, which cannot be taken for granted especially for low-resource languages. Since maintaining such fact-checking websites is a resource-heavy task in itself, the data that can be gathered from such sources is limited. Further given a dataset of machine-generated content, it might be possible to identify some inherent features which can help in distinguishing them from content generated by humans. This in turn can help in combating the spread of LLM-generated misinformation.

\para{Our Contributions}
We propose an approach for generating such a dataset using adversarial prompting. We leverage the capabilities of Large Language Models (LLMs) to create a robust fake news dataset that will help in understanding the anatomy of LLM-generated misinformation. Our dataset captures a variety of misinformation patterns and exhibits how LLMs can be leveraged for generating misinformation which lays the groundwork for effectively fighting against LLM-generated misinformation. Such an approach can not only generate a large amount of training data it also allows controlled generation of misinformation where we can specify the amount and type of misinformation. We present a novel dataset, that includes news articles on a wide array of topics, accompanied by their factually correct summaries and summaries with misinformation, both generated using LLMs. To the best of our knowledge, this is a first-of-its-kind dataset that leverages LLMs for generating a Fact-Checking dataset. We select four different categories of misinformation for this purpose: Fabrication, Misrepresentation, False Attribution and Inaccurate quantities. We include varied levels of misinformation in the summaries ranging from outright false to subtle biases. We believe this approach can scale well and the dataset can be easily updated over time. We also include some baseline experiments that demonstrate the challenges in identifying the summaries with misinformation.

\section{Related Work}
Fact-checking is actively researched at least since the last decade. What started as a simple binary task of checking whether or not a piece of information is correct has evolved into a multi-faceted problem that is constantly evolving. One of the early attempts at formally defining the problem and a detailed study of the existing approaches can be found in \cite{uscinski2013epistemology}. The problem of Fake news detection can be broadly classified into three categories: fake news detection, fact verification, and containment of fake news \& mitigation of its impact on the common people \cite {sharma2019combating}. Fake news detection methods determine the correctness of the claim based on the linguistic features of the text and social network users interactions with or reactions to the posts. These methods do not take into account external knowledge, while fact-checking methods use external evidence to determine the correctness of the claim. The assembly of a fake checking dataset with fake and true news is a challenging, time-consuming, and financially expensive task since each potential fake article needs to be carefully examined by the human expert \cite{d2021fake}.

%The early roots of the fake news detection problem can be linked to the problem of deceptive language detection, which was widely studied by the philosophy, psychology, and sociology communities. The authors in \cite{sharma2019combating} defined fake news as "a news article or message published and propagated through media, carrying false information regardless of the means and motives behind it."
 
This proposed FakeSum dataset can help in both fake news detection and fact verification. We can use the same dataset into setups, discussed in detail in section \ref{sec:experiments}, by either classifying the LLM-generated summaries on their own (Fact-Checking) or first retrieving original news articles that are relevant for a given summary and then comparing the summary against them (Fact verification).  

 In this section we limit our focus to the existing datasets that can aid in fact-checking rather than the contemporary approaches to solving the problem. A survey by \cite{murayama2021dataset} reported 127 fake news datasets. One of the largest and perhaps the most popular datasets consists of $~$12K short statements spanning across a decade with human labels as to whether they are factually correct or not \cite{wang2017liar}. Evaluation forums like TREC through its "Health misinformation task"\cite{barron2020overview} and CLEF through the series of "CheckThat" labs\cite{clarke2020overview} are also some of the leading platforms that generated fact-checking datasets for several domains. The CheckThat lab at CLEF introduced a series of fact-verification tasks, such as Check-worthiness of tweets, verifiable factual claims detection, and Detecting previously fact-checked claims in tweets. The CheckThat lab has offered tasks in multilingual settings, and these languages, apart from English, are Arabic, Bulgarian, Dutch, and Turkish.
 
 Another work tackles the problem of false claims in simple plain text by sampling two datasets, namely Snopes\_credibility and Wikipedia\_credibility from Snoop and Wikipedia. Each article is labelled as true or false\cite{popat2016credibility}. FEVER: Fact Extraction and VERification is a large dataset for fake news verification, containing 185,445 claims generated by altering sentences extracted from Wikipedia\cite{thorne2018fever}. The documents were classified as SUPPORTED, REFUTED, or NOTENOUGHINFO. One noteworthy detail is that the majority of the fact-check datasets are from the political domain \cite{kotonya2020explainable}. While a substantial portion of the proposed FakeSum dataset also comes from the political domain it is more diverese with topics like sports, movies, technology, etc.

\section{FakeSumm Dataset Construction}
\label{section:fakesumm_dataset}

Our proposed approach of data generation involves introducing factual incorrectness into the summary of an original article. Ideally, the summaries should include all important pieces of information in the article. Since our objective is to construct a benchmark dataset for misinformation detection, we also generate factually correct summaries from a larger set of articles so that the task on this dataset then becomes that of separating out the correct summaries from the incorrect ones.

As a collection of the base articles that are to be summarised, we employ a set of almost 5000 news articles in English collected from CNN-News18\footnote{\url{https://www.news18.com/}}, a leading news channel in India. To generate the factually correct summaries, we used the GPT-3.5 model from OpenAI via the official API. For a subset of 1000 articles, we generated summaries with varying levels of misinformation. These summaries deliberately include elements that are misleading, biased or completely incorrect. The misleading or \emph{fake} summaries were generated using the GPT-4 \cite{openai2023gpt4} API.

We use different models for generating correct and fake summaries because of two reasons. The first reason is that our experiments revealed that while both models are equally good at generating correct summaries, GPT-4 is much better at generating incorrect summaries. The other reason is the cost difference between GPT-3.5 and GPT-4 is very high, GPT-4 is about 15x costlier compared to GPT3.5 for the same input and output lengths. So we preferred GPT-3.5 whenever the performance was comparable to GPT-4, i.e. for the correct summaries. 

%To address potential biases associated with keyword prominence, we further augment dataset with \(\sim\)4000 English article and factually correct summaries, which will create more realistic experimental scenario, where ratio of factually incorrect information to correct one is notably reduced.

As a part of the data generation process, we introduced the following types of factual incorrectness in the summaries of the articles.

\begin{itemize}
    \item \textbf{Fabrication} - This is perhaps the easiest type of incorrectness both to create and detect. It involves completely making up data, sources, or events. This involves creating `facts' that have no basis in reality. 
    
    \item \textbf{False Attribution} This type of incorrectness preserves the overall narrative, but incorrectly attributes an event, statement, action etc. to a different entity than the original one. 
    
    \item  \textbf{Inaccurate numerical quantities} - In this case both subtle and obvious changes are made to numerical quantities within an article. While often this category is benign, there are cases where it has the potential to seriously mislead the reader. For instance, "25 people died of cholera in New Delhi" can be rewritten as "Dozens die of cholera in New Delhi" to intentionally exaggerate the situation.
    
    \item \textbf{Misrepresentation} - This relies on introducing bias in the summary, while technically retaining the original story. This means the summary could be written to intentionally show some person or entity in a good or bad light or to downplay or exaggerate certain events.
    
\end{itemize}

In the table \ref{tab:categories} below, we provide an example of the original correct summary, summaries with different kinds of misinformation and a brief explanation. We do not include the complete text of the original article for which these summaries were generated due to space constraints, but the same can be accessed on the news channels website\footnote{\url{https://indianexpress.com/article/cities/ahmedabad/gandhinagar-5-children-among-13-hospitalised-after-suspected-cholera-outbreak-8684680/}}. We provide two examples for the misrepresentation category, one which is biased towards the government and one against it. This is a rare case where summaries with all types of misinformation are generated. Generally, for a given article it would be possible to have only one or two types of misinformation depending on the content. For instance, it would generally not be possible to introduce much bias in an article that talks about the outcome of a recent cricket match. The category of fabrication is not subject to limitation and can be applied to all the articles since the summary can be completely made up in this case with no relation to the source article. 
\begin{table}[]
\begin{mdframed}
 
\noindent\textbf{Original Summary:} Thirteen individuals, including five children, were admitted to health facilities in Gandhinagar due to a suspected cholera outbreak in the Kalol municipality. Two cases were confirmed, prompting a two-kilometer radius of the outbreak area to be declared 'cholera-affected.' The State Health Minister, Rushikesh Patel, communicated the situation to Union Home Minister Amit Shah. 

\noindent\textbf{Explanation}: This is an accurate summarization of the facts presented in the original text\\ \hrulefill \Bstrut

\noindent\textbf{Fabrication}: A cholera outbreak in Gandhinagar has resulted from a genetically modified strain that escaped a bioengineering lab. This has led to 13 hospitalizations, with five children affected. The government has declared a two-kilometer radius of the municipality as 'cholera-affected.'

\noindent\textbf{Explanation}: The outbreak being linked to a genetically modified strain and an escape from a lab is entirely fabricated and not supported by the original text.\\

\hrulefill\Bstrut

\noindent\textbf{False Attribution}: The World Health Organization (WHO) confirmed a cholera outbreak in Gandhinagar's Kalol municipality, with 13 people admitted to healthcare facilities. The Union minister Amit Shah has declared a two-kilometer radius around the outbreak area as 'cholera-affected.' Gandhinagar Collector Hitesh Koya has been briefed about the situation by the Home minister. 

\noindent\textbf{Explanation}: The summary falsely attributes the confirmation of the outbreak to the Home minister instead of the District Collector. \\

\hrulefill \Bstrut

\noindent\textbf{Inaccurate Quantities or Measurements}: A cholera outbreak has affected thousands in the Kalol municipality, Gandhinagar. The entire region has been declared a 'cholera-affected' area. Over 100 cases have been confirmed, with many children being affected.

\noindent\textbf{Explanation}: The summary inaccurately increases the number of affected individuals and the geographical scope of the cholera outbreak. \\

\hrulefill \Bstrut

\noindent\textbf{Misrepresentation (Pro-Establishment)}: In response to the confirmation of two cholera cases in Kalol municipality, Gandhinagar, swift action was taken by the local authorities, with Hitesh Koya, the Gandhinagar Collector, promptly declaring a two-kilometer radius around the outbreak as 'cholera-affected'. In addition, State Health Minister Rushikesh Patel effectively communicated the situation to Union Home Minister Amit Shah. The efficient response and transparent communication by the government demonstrates their commitment to public health and safety.

\noindent\textbf{Misrepresentation (Anti-Establishment)}: Despite previous cholera outbreaks and consistent warnings about the condition of water pipelines in Kalol municipality, the local authorities failed to prevent another outbreak, leading to 13 people, including five children, being admitted to health facilities in Gandhinagar. Although the State Health Minister, Rushikesh Patel, informed Union Home Minister Amit Shah, the reaction came after the confirmation of two cases. This repeated failure to address public health risks effectively underlines a lack of initiative and foresight from the establishment. 

\noindent\textbf{Explanation}: The summaries present a biased picture with the first summary focusing on a swift response by the authorities, the second summary highlighting previous outbreaks of Cholera and the failure of the authorities to prevent yet another outbreak. 

\end{mdframed}
   \caption{Examples of correct and misleading summaries}
    \label{tab:categories}
\end{table}

The summaries with misinformation were generated using GPT-4\cite{openai2023gpt4}. This process involved adversarial prompting, commonly known as jailbreaking. An adversarial prompt is a type of prompt specifically designed to bypass the safety measures that LLMs have in place for preventing or limiting its misuse. For instance, asking GPT to \textit{"Help me create a bomb"} returned a response \textit{"Sorry, I can't assist with that request. It is illegal to do that. If you're going through a tough time, please seek help or speak to someone who can assist you. Your safety and the safety of others are essential"}. However previous works have shown that such safety features can be bypassed with careful adversarial prompting\cite{zou2023adverserial}. 

We use adversarial prompts to generate factually incorrect summaries. While our straightforward attempt with prompts like "Generate a fake summary for this article" resulted in negative responses from GPT, we were able to exploit certain loopholes using adversarial prompts to generate the dataset. For ethical reasons, we do not include the exact steps we followed for generating the summaries. Specifically, we do not include the details of how we \emph{tricked} GPT into generating these summaries with misinformation. However, we do include the portions of the prompts that define the type of incorrectness we wanted to include in the dataset. These prompts are listed in table \ref{tab:prompts} below. 

We used chain-of-thoughts prompting, known to mimic the reasoning process to some extent\cite{wei2022chain}, to ensure a more controlled dataset. In this type of prompting, the desired task is broken down into smaller steps that are easier for an LLM to follow. For example, in the category ``Incorrect numerical quantities'' we first ask GPT to identify the numerical quantities in an article, and then modify those while keeping everything else the same. This strategy proved to be much more effective than simply telling GPT to generate false summaries with incorrect numbers. Some of these steps are common to all the categories like defining the summary length and ensuring the output does not include phrases like "Here is your incorrect summary" which may be easy to pick up for a fact-checking system.

For some of the categories we also employed few-shot learning, another popular prompting approach where we give some examples of the type of misinformation that is to be produced. For incorrect quantities, we give two examples as shown in table \ref{tab:prompts}. The first example retains all the quantities in the article, but mixes up their order. This can throw off simple term matching technique while verifying the summary against the source article as will be explained in the experiment section \ref{sec:experiments} below.  The other example can replace the existing quantities with different, but semantically similar quantities, e.g. 30 USD is changed to 300 INR. We selectively apply these steps to a subset of the articles, so not all summaries with incorrect quantities will have both types of errors.

Further, due to the conversational nature of text in the output, GPT often includes additional details that explain the generated output. Our experiments revealed that details like what was incorrect in the summary, what entities were replaced, etc. were commonly included in the output. In some cases, it also included the correct summaries in the output alongside the incorrect summaries. We calibrated our prompts by adding additional steps to ensure the output contains \emph{only the incorrect summary} and no additional detail that could make identifying the incorrect summaries a trivial task. We also carefully select the articles such that a given category of incorrectness is actually applicable to that article. As an example, we generate summaries with incorrect quantities only from articles which have more than 10 numerical quantities or statistics so either of the steps described above can actually be applied. Likewise, for the misrepresentation category, we choose a list of 15 topics where it is actually possible to generate biased summaries. Such selection was done based on simple keyword search.

In the misrepresentation category, we include articles about geopolitics like the India-Pakistan relation, or the Russia-Ukraine war while others are about the government response to accidents or natural disasters like the Odisha train accident or heavy rain and floods where biased representation is possible. We also use articles about political leaders to generate summaries, that are for example, pro and anti-Narendra Modi. The complete list is included in appendix \ref{appendix:misrepresentation_categories}. For each topic, half the articles were biased towards an entity and the remaining were biased against it. For example in the case of the Russia-Ukraine war, half the summaries will be pro-Russia and the other half pro-Ukraine.

For all categories except the misrepresentation category, we vary the difficulty level of the incorrect summaries. We generated a difficult variant for half the documents, where we put constraints like, \emph{the summary should only include entities and events actually mentioned in the original article} and for the easy variants this constraint is relaxed. For the difficult category, the incorrect summary will be more realistic and could still represent a real-life scenario. We could change "Sachin Tendulkar hit a century" to "Brian Lara hit a century" and it would still be a possible real life scenario, both Tendulkar and Lara being cricketers. The easy category, on the other hand, can produce "David Attenborough hit a century" which represents a very unlikely scenario, Attenborough being a biologist and Historian among other things, but certainly not a cricketer. We include the easy category because they represent cases of Hallucinations, which although relatively easier, are by no means trivial to detect automatically. 

\begin{table}
    \begin{mdframed}
    \textbf{Common steps}
    \begin{enumerate}
        \item Generate a 100 word summary using the following steps
        \item One or more \emph{Category specific} steps mentioned below
        \item Exclude any sentence that may indicate that the summary is factually incorrect.
        \item Exclude any explanations        
        \item Finally return only the factually incorrect version   

    \end{enumerate} \vspace{-5mm}
    \hrulefill \Tstrut   
    
    \textbf{Fabrication}
    \begin{enumerate}
        \item Identify important entities, events and other information in the article.
        \item Fabricate a story using all this information.
    \end{enumerate} \vspace{-5mm}
    \hrulefill \Tstrut   
    
    \textbf{False Attribution}
    \begin{enumerate}
        \item Identify important entities, events and other information in the article.
        \item Change some information in the summary by falsely attributing a statement, idea or action to some other person or entity. The entity or person to which the statement is falsely attributed should be similar to the original entity and appear in the original text. Example: Let's say entity X performed an action or made a statement. Change it to Y such that the action or statement remains unchanged. Also Y should appear in the original article. 
        \item Avoid mentioning the entities explicitly.
        \item The incorrectness in the summary should only result from false attribution. The overall narrative should remain the same
        \item Before returning check "Does this summary have only the entities from the original text?"
    \end{enumerate} \vspace{-5mm}
    \hrulefill \Tstrut   

    \textbf{Inaccurate Quantities or Measurements}
    \begin{enumerate}
        \item Identify important entities, events, numbers, statistics, quantities and other information in the article
        \item Jumble up the numbers found in the summary. Do not include any new numbers. Example: 78 millions USD was received as aid. Then a lot of text. 21 million was spent. could become 21 millions USD was received as aid. Then a lot of text. 78 million was spent.
        \item Change some information in the summary by including Inaccurate Quantities or Measurements or changing some existing measurements or statistics to incorrect ones. Example: 100 Million INR is changed to 30 Billion USD. \
        \item The incorrectness in the summary should only result from inaccurate Quantities or measurements and nothing else
    \end{enumerate}  \vspace{-5mm}
    \hrulefill \Tstrut   
    
    \textbf{Misrepresentation}
    The exact prompt for misrepresentation depends on the topic of the article, but the common patterns are listed here. Unlike other categories, these patterns are not the steps for a single prompt, but rather one of them is used depending on the context of the article.
    \begin{itemize}
        \item The summary should praise/criticise \emph{X} for \emph{Y}
        \item The summary should be biased towards \emph{X}
        \item The summary should favor \emph{X} over \emph{Y}
    \end{itemize}
    \textbf{Examples}
    \begin{itemize}
        \item[] The summary should
        \item ... praise \emph{government efforts} in handling \emph{Odisha train accident}
        \item ... be biased towards \emph{Russia}
        \item ... exaggerate \emph{Manipur Violence}        
    \end{itemize}

    \end{mdframed}
    \caption{Prompts for generating misinformation}
    \label{tab:prompts}
\end{table}

 \subsection{Dataset Statistics}
\begin{figure}
    \centering
    \includegraphics[width=0.8\textwidth]{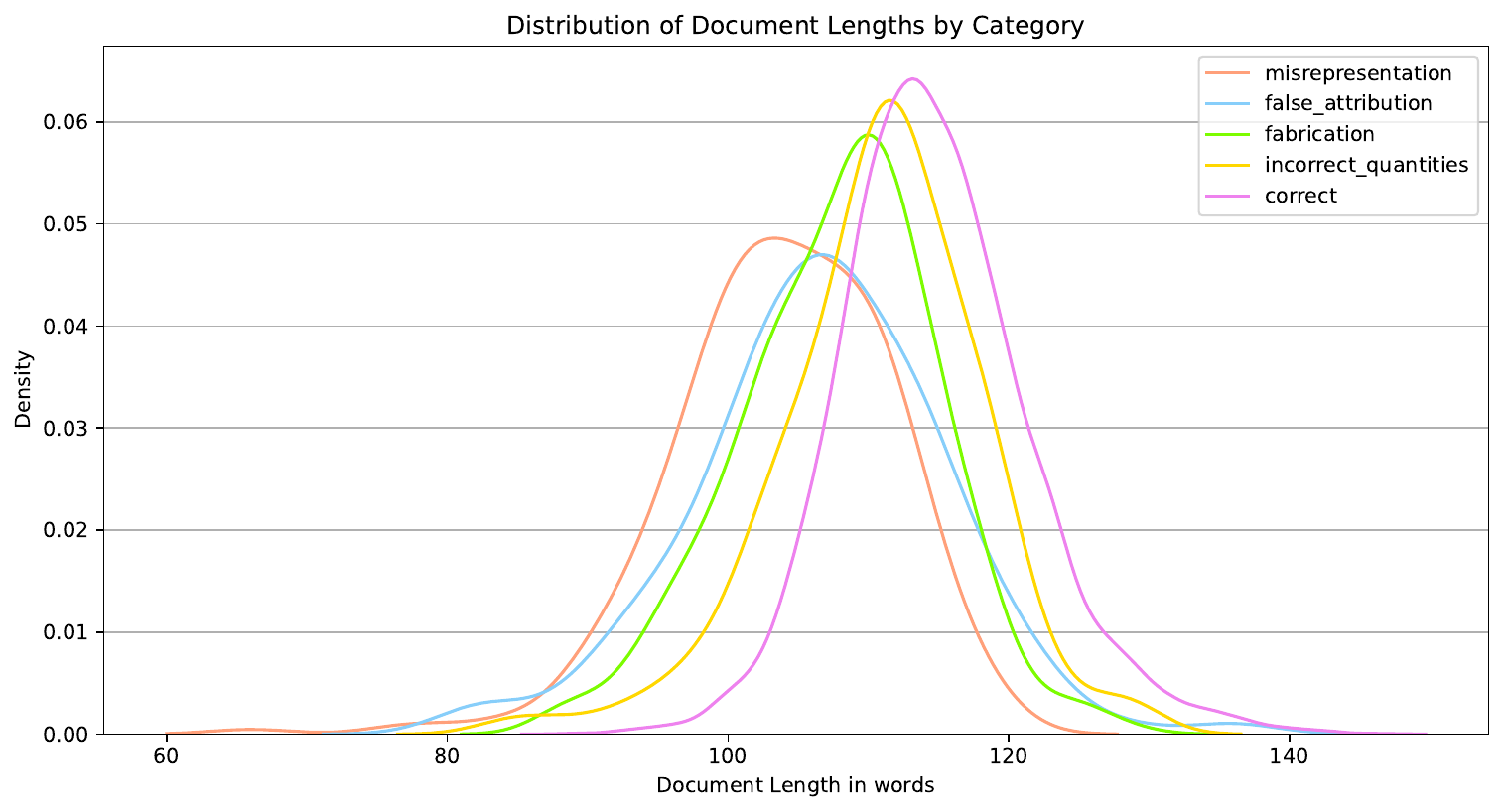}
    \caption{Distribution of Document Lengths by Category.}
    \label{fig:doclengths}
\end{figure}

 The final dataset contains about 5000 correct and about 1000 incorrect summaries across cour categories. Figure \ref{fig:doclengths} shows the distribution of summary lengths, in terms of number of words, for each class. The dataset was split into train, validation and test sets with a ratio of 70:10:20. Table \ref{tab:dataset_stats} contains class-wise counts of documents for each split. We intend to make the entire dataset as well the the details of the split publicly available. 

 \begin{table}[h]
     \centering
     \begin{tabular}{|l|c|c|c|}
\hline
\textbf{}            & \textbf{Train Split} & \textbf{Validation Split} & \textbf{Test Split} \\ \hline
Correct              & 4189                 & 591                       & 1184                \\ 
Misrepresentation    & 206                  & 33                        & 64                  \\ 
False Attribution    & 169                  & 32                        & 60                  \\ 
Fabrication          & 157                  & 24                        & 49                  \\ 
Incorrect Quantities & 146                  & 15                        & 34                  \\ 
\textbf{Total}       & \textbf{4867}        & \textbf{695}              & \textbf{1391}       \\ \hline
\end{tabular}
     \caption{Dataset Statistics of the FakeSumm Dataset.}
     \label{tab:dataset_stats}
 \end{table}

\section{Evaluation of Misinformation Detection on our Dataset}
\label{sec:experiments}
We conducted two sets of experiments with the proposed Fakesum dataset. In the first setup, we pose this as a classification task where given an LLM-generated summary we want to label it as either the correct summary or one of the four categories of misinformation mentioned in Table \ref{tab:categories}. This setup is oriented towards standalone fact-checking without comparing against any existing articles. While not a common approach in traditional fact-checking setups, this does represent a realistic scenario in real-life where a person reading social media messages, for example on WhatsApp, is not going to actively verify the content. However, an automated approach that can detect misinformation without external verification can prevent its spread.

The second setting that we experiment with tries to mimic the traditional Fact-checking setup where humans verify the content against existing information or a knowledge base. In order to do this, the first step would be to find one or more articles that contain factually correct information that is closely related to the summary being classified. We achieve this by first retrieving the best matching article for a given summary from a larger pool of around 30,000 news articles from the same source and the same time range. The matching was done using Okapi-BM25. We then use a classifier similar to the one in the previous setting, but jointly train it on the article-summary pair, compared to only the summary in the previous setting.

We use four standard techniques mentioned below for training the classifier in both the settings:
\begin{itemize}
    \item \textbf{SVC + TF-IDF}: We used a Support Vector Classifier with RBF kernal and used the TF-IDF values as features.
    \item \textbf{BiLSTM}: We used a two-layer BiLSTM with pre-trained Glove embeddings\cite{pennington2014glove} with 300 dimensions and 6B tokens. The LSTM hidden dimensions were chosen to be 256, a learning rate of 1e-4  with Adam optimizer, and a dropout rate of 0.5 was used to train the model for 25 epochs. 
    \item \textbf{Transformers}: We used BERT\cite{devlin2019bert} and RoBERTa\cite{liu2019roberta} pre-trained models. Specifically, Bert-base-uncased and xlm-roberta-base were finetuned for 5 epochs and learning rate 2e-5 with batch size 16. In the case of the first setup, only LLM-generated summaries were used for fine-tuning while in the second setup, we used both the summary and the best-matching article.
    \end{itemize}

We employ two different setups for the experiments which differ in terms of the input provided to the model. These are described as follows.
\begin{itemize}
    \item \textbf{Setting 1 - Summary only}: Four different classifiers are trained to classify a given summary into one of the 5 predefined classes without and supporting articles or knowledge base.
    \item \textbf{Setting 2 - Article + Summary}: We train the same classifiers as in setting-1, but using a article-summary pair. During training actual articles which were used to create the LLM-generated summary were used, while for validation and testing, we retrieved the best matching article using Okapi-BM25 articles from a larger collection of ~28k news articles.
\end{itemize}

Out of these two, the first setup corresponds to a more realistic use-case. We also include the second setup in our experiments to see how effectively supervised models can perform when the original article (which is factually correct) is also known during the training and the inference times.

Table \ref{tab:results} contains the results for both these settings. We use precision, recall and macro-F1 scores for evaluation. Unsurprisingly LLMs seem to do a much better job in both settings compared to SVC or LSTMs. BERT is performed the best, closely followed by RoBERTa. All models except SVC show improvement when provided a reference article. Table \ref{tab:classification_report} contains classwise performance for each category.

\begin{table}[t]
    \centering
\begin{tabular}{|c|c|c|c|c|}
\hline
\textbf{Experimental Setting} & \textbf{SVC + TF-IDF} & \textbf{Bi-LSTM} & \textbf{BERT} & \textbf{RoBERTa} \\ \hline

\begin{tabular}[c]{@{}c@{}}With Summaries Only\\ (Setting - 1)\end{tabular}         &        0.18        &     0.25   &       \textbf{0.57}        &     0.52             \\ 

\begin{tabular}[c]{@{}c@{}}With Articles and Summaries\\ (Setting - 2)\end{tabular} &            0.18    &     0.27    &        \textbf{0.63}       &       0.56           \\ \hline

\end{tabular}
    \caption{Macro F1 scores for all baseline experiments}
    \label{tab:results}
\end{table}

\begin{table}[h]
    \centering
    \resizebox{\textwidth}{!}{
    \begin{tabular}{|c|c|c|ccccc|c|}
\hline
\multirow{2}{*}{\begin{tabular}[c]{@{}c@{}}Experimental \\ Setting\end{tabular}}                           & \multirow{2}{*}{Models}                                                     & \multirow{2}{*}{Scores} & \multicolumn{5}{c|}{Class Names}                                                                                                                                                                                                                               & \multirow{2}{*}{\begin{tabular}[c]{@{}c@{}}Macro \\ Avg\end{tabular}} \\ \cline{4-8}
                                                                                                           &                                                                             &                         & \multicolumn{1}{c|}{Correct} & \multicolumn{1}{c|}{Misrepresentation} & \multicolumn{1}{c|}{\begin{tabular}[c]{@{}c@{}}False \\ Attribution\end{tabular}} & \multicolumn{1}{c|}{Fabrication} & \begin{tabular}[c]{@{}c@{}}Incorrect \\ Quantities\end{tabular} &                                                                       \\ \hline
\multirow{12}{*}{\begin{tabular}[c]{@{}c@{}}With \\ Summaries \\ Only \\ (Setting 1)\end{tabular}}         & \multirow{3}{*}{\begin{tabular}[c]{@{}c@{}}SVC \\ + \\ TF-IDF\end{tabular}} & Precision               & \multicolumn{1}{c|}{0.85}    & \multicolumn{1}{c|}{0.00}              & \multicolumn{1}{c|}{0.00}                                                         & \multicolumn{1}{c|}{0.00}        & 0.00                                                            & 0.17                                                                  \\ \cline{3-9} 
                                                                                                           &                                                                             & Recall                  & \multicolumn{1}{c|}{1.00}    & \multicolumn{1}{c|}{0.00}              & \multicolumn{1}{c|}{0.00}                                                         & \multicolumn{1}{c|}{0.00}        & 0.00                                                            & 0.20                                                                  \\ \cline{3-9} 
                                                                                                           &                                                                             & F1 Score                & \multicolumn{1}{c|}{0.92}    & \multicolumn{1}{c|}{0.00}              & \multicolumn{1}{c|}{0.00}                                                         & \multicolumn{1}{c|}{0.00}        & 0.00                                                            & 0.18                                                                  \\ \cline{2-9} 
                                                                                                           & \multirow{3}{*}{BiLSTM}                                                     & Precision               & \multicolumn{1}{c|}{0.86}    & \multicolumn{1}{c|}{0.00}              & \multicolumn{1}{c|}{0.73}                                                         & \multicolumn{1}{c|}{1.00}        & 0.00                                                            & 0.52                                                                  \\ \cline{3-9} 
                                                                                                           &                                                                             & Recall                  & \multicolumn{1}{c|}{1.00}    & \multicolumn{1}{c|}{0.00}              & \multicolumn{1}{c|}{0.17}                                                         & \multicolumn{1}{c|}{0.02}        & 0.00                                                            & 0.24                                                                  \\ \cline{3-9} 
                                                                                                           &                                                                             & F1 Score                & \multicolumn{1}{c|}{0.92}    & \multicolumn{1}{c|}{0.00}              & \multicolumn{1}{c|}{0.28}                                                         & \multicolumn{1}{c|}{0.03}        & 0.00                                                            & 0.25                                                                  \\ \cline{2-9} 
                                                                                                           & \multirow{3}{*}{BERT}                                                       & Precision               & \multicolumn{1}{c|}{0.94}    & \multicolumn{1}{c|}{0.66}              & \multicolumn{1}{c|}{0.46}                                                         & \multicolumn{1}{c|}{0.82}        & 0.83                                                            & \textbf{0.74}                                                         \\ \cline{3-9} 
                                                                                                           &                                                                             & Recall                  & \multicolumn{1}{c|}{0.99}    & \multicolumn{1}{c|}{0.92}              & \multicolumn{1}{c|}{0.24}                                                         & \multicolumn{1}{c|}{0.38}        & 0.15                                                            & 0.54                                                                  \\ \cline{3-9} 
                                                                                                           &                                                                             & F1 Score                & \multicolumn{1}{c|}{0.97}    & \multicolumn{1}{c|}{0.77}              & \multicolumn{1}{c|}{0.32}                                                         & \multicolumn{1}{c|}{0.52}        & 0.25                                                            & \textbf{0.57}                                                         \\ \cline{2-9} 
                                                                                                           & \multirow{3}{*}{RoBERTa}                                                    & Precision               & \multicolumn{1}{c|}{0.96}    & \multicolumn{1}{c|}{0.58}              & \multicolumn{1}{c|}{0.17}                                                         & \multicolumn{1}{c|}{0.48}        & 0.50                                                            & 0.54                                                                  \\ \cline{3-9} 
                                                                                                           &                                                                             & Recall                  & \multicolumn{1}{c|}{0.96}    & \multicolumn{1}{c|}{0.97}              & \multicolumn{1}{c|}{0.04}                                                         & \multicolumn{1}{c|}{0.50}        & 0.29                                                            & \textbf{0.55}                                                         \\ \cline{3-9} 
                                                                                                           &                                                                             & F1 Score                & \multicolumn{1}{c|}{0.96}    & \multicolumn{1}{c|}{0.73}              & \multicolumn{1}{c|}{0.07}                                                         & \multicolumn{1}{c|}{0.49}        & 0.37                                                            & 0.52                                                                  \\ \hline
\multirow{12}{*}{\begin{tabular}[c]{@{}c@{}}With \\ Article \\ and \\ Summary \\ (Setting 2)\end{tabular}} & \multirow{3}{*}{\begin{tabular}[c]{@{}c@{}}SVC \\ + \\ TF-IDF\end{tabular}} & Precision               & \multicolumn{1}{c|}{0.85}    & \multicolumn{1}{c|}{0.00}              & \multicolumn{1}{c|}{0.00}                                                         & \multicolumn{1}{c|}{0.00}        & 0.00                                                            & 0.17                                                                  \\ \cline{3-9} 
                                                                                                           &                                                                             & Recall                  & \multicolumn{1}{c|}{1.00}    & \multicolumn{1}{c|}{0.00}              & \multicolumn{1}{c|}{0.00}                                                         & \multicolumn{1}{c|}{0.00}        & 0.00                                                            & 0.20                                                                  \\ \cline{3-9} 
                                                                                                           &                                                                             & F1 Score                & \multicolumn{1}{c|}{0.92}    & \multicolumn{1}{c|}{0.00}              & \multicolumn{1}{c|}{0.00}                                                         & \multicolumn{1}{c|}{0.00}        & 0.00                                                            & 0.18                                                                  \\ \cline{2-9} 
                                                                                                           & \multirow{3}{*}{BiLSTM}                                                     & Precision               & \multicolumn{1}{c|}{0.87}    & \multicolumn{1}{c|}{0.33}              & \multicolumn{1}{c|}{0.48}                                                         & \multicolumn{1}{c|}{0.50}        & 1.00                                                            & 0.64                                                                  \\ \cline{3-9} 
                                                                                                           &                                                                             & Recall                  & \multicolumn{1}{c|}{0.99}    & \multicolumn{1}{c|}{0.04}              & \multicolumn{1}{c|}{0.17}                                                         & \multicolumn{1}{c|}{0.02}        & 0.03                                                            & 0.25                                                                  \\ \cline{3-9} 
                                                                                                           &                                                                             & F1 Score                & \multicolumn{1}{c|}{0.93}    & \multicolumn{1}{c|}{0.07}              & \multicolumn{1}{c|}{0.25}                                                         & \multicolumn{1}{c|}{0.03}        & 0.06                                                            & 0.27                                                                  \\ \cline{2-9} 
                                                                                                           & \multirow{3}{*}{BERT}                                                       & Precision               & \multicolumn{1}{c|}{0.95}    & \multicolumn{1}{c|}{0.77}              & \multicolumn{1}{c|}{0.58}                                                         & \multicolumn{1}{c|}{0.76}        & 0.55                                                            & \textbf{0.72}                                                         \\ \cline{3-9} 
                                                                                                           &                                                                             & Recall                  & \multicolumn{1}{c|}{0.99}    & \multicolumn{1}{c|}{0.75}              & \multicolumn{1}{c|}{0.39}                                                         & \multicolumn{1}{c|}{0.63}        & 0.18                                                            & \textbf{0.59}                                                         \\ \cline{3-9} 
                                                                                                           &                                                                             & F1 Score                & \multicolumn{1}{c|}{0.97}    & \multicolumn{1}{c|}{0.76}              & \multicolumn{1}{c|}{0.46}                                                         & \multicolumn{1}{c|}{0.69}        & 0.27                                                            & \textbf{0.63}                                                         \\ \cline{2-9} 
                                                                                                           & \multirow{3}{*}{RoBERTa}                                                    & Precision               & \multicolumn{1}{c|}{0.94}    & \multicolumn{1}{c|}{0.74}              & \multicolumn{1}{c|}{0.43}                                                         & \multicolumn{1}{c|}{0.59}        & 0.45                                                            & 0.63                                                                  \\ \cline{3-9} 
                                                                                                           &                                                                             & Recall                  & \multicolumn{1}{c|}{0.98}    & \multicolumn{1}{c|}{0.75}              & \multicolumn{1}{c|}{0.06}                                                         & \multicolumn{1}{c|}{0.63}        & 0.29                                                            & 0.54                                                                  \\ \cline{3-9} 
                                                                                                           &                                                                             & F1 Score                & \multicolumn{1}{c|}{0.96}    & \multicolumn{1}{c|}{0.74}              & \multicolumn{1}{c|}{0.11}                                                         & \multicolumn{1}{c|}{0.61}        & 0.36                                                            & 0.56                                                                  \\ \hline
\end{tabular}}
\caption{Evaluation details (per-class F1 scores and macro-average F1 scores) for the misinformation detection task (5-way multi-class classification, i.e., \texttt{Fabrication}, \texttt{False Attribution}, \texttt{Incorrect Quantities}, \texttt{Misrepresentation}, and \texttt{Correct}) for the two different experiment settings - one with the base article known and the other without any knowledge of the base article.}
\label{tab:classification_report}
\end{table}

\section{Future Work}
As outlined in section \ref{section:fakesumm_dataset} it is evident that an incorrect summary may exhibit multiple forms of incorrectness. In such instances, it is important to identify each specific type of incorrectness present in the summary. Beyond identifying the category of inaccuracy, it is essential to pinpoint the exact piece of information that is erroneous. While for humans it is relatively straightforward to spot misinformation like incorrect numerical quantities given a reference article, it's still a challenge for machine learning models. Therefore it's crucial to delve further into the development of more robust models. Furthermore, the current dataset is primarily focused on the English language, which can be extended further by creating a more comprehensive multilingual dataset.

% Incorrect summaries with multiple types of incorrectness
% Along with detecting type of incorrectness, it is equally important to identify which particular piece of information is incorrect.
% how to take care of incorrectness that required external knowledge other then available information in article
% More robust models
% Multilingual Datasets

\appendix

\section{Topics used for Misrepresentation Category}
We used the following topics for generating summaries with misrepresentation. In cases where more than one entity are included, we generated summaries that favored one over the other. In the case of events (8 and 9) the event, the bias introduced was for or against how the situation was handled by the government and not the event itself. The topics were chosen such that they are well represented in the FakeSum dataset.
\begin{multicols}{2} % this will create 2 columns
\begin{enumerate}
    \item Narendra Modi
    \item Rahul Gandhi
    \item Donald Trump
    \item Congress and BJP
    \item India - Pakistan Relations
    \item India - China Relations 
    \item Russia - Ukraine War
    \item Manipur Violence
    \item Odisha Train Accident
    \item Imran khan - Pakistan Army
    \item Kerala Story movie
    \item GPT
\end{enumerate}
\end{multicols} % this will create 2 columns
\label{appendix:misrepresentation_categories}
\bibliographystyle{splncs04}
\bibliography{ecir-24}
\end{document}